\documentclass[10pt,twocolumn,letterpaper,table]{article}
\usepackage[pagenumbers]{cvpr} %
\usepackage{graphicx}
\usepackage{amsmath}
\usepackage{amssymb}
\usepackage{times}
\usepackage{epsfig}
\usepackage{caption}
\usepackage{subcaption}
\usepackage{tabularx}
\usepackage{booktabs}
\usepackage{bbding}
\usepackage{graphics}
\usepackage{array}
\usepackage[accsupp]{axessibility}
\usepackage{multirow}
\usepackage{algorithm}
\usepackage{algorithmic}

\definecolor{cvprblue}{rgb}{0.21,0.49,0.74}
\usepackage[pagebackref,breaklinks,colorlinks,allcolors=cvprblue]{hyperref}
\newcommand{\symbolHt}{1.1em}

\newcommand{\locChar}{%
\centering
  \begingroup\normalfont
  \raisebox{-0.18\height}
  {\includegraphics[height=\symbolHt]{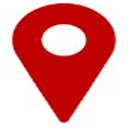}}%
  \endgroup
}

\newcommand{\textChar}{%
\centering
  \begingroup\normalfont
  \raisebox{-0.18\height}
  {\includegraphics[height=\symbolHt]{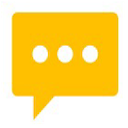}}%
  \endgroup
}

\newcommand{\contChar}{%
  \begingroup\normalfont
  \raisebox{-0.18\height}{\includegraphics[height=\symbolHt]{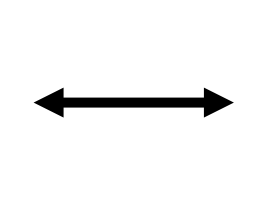}}%
  \endgroup
}
\newcommand{\satChar}{%
  \begingroup\normalfont
  \raisebox{-0.18\height}{\includegraphics[height=\symbolHt]{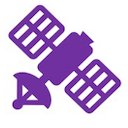}}%
  \endgroup
}
\newcommand{\plusChar}{%
  \begingroup\normalfont
  \raisebox{-0.18\height}
  {\includegraphics[height=\symbolHt]{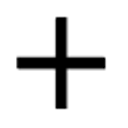}}%
  \endgroup
}
\newcommand{\arrowChar}{%
  \begingroup\normalfont
  \raisebox{-0.18\height}
  {\includegraphics[height=\symbolHt]{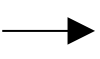}}%
  \endgroup
}
\newcommand{\imageChar}{%
\centering
  \begingroup\normalfont
  \raisebox{-0.18\height}
  {\includegraphics[height=\symbolHt]{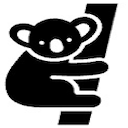}}%
  \endgroup
}
\newcommand{\audioChar}{%
  \begingroup\normalfont
  \raisebox{-0.18\height}
  {\includegraphics[height=\symbolHt]{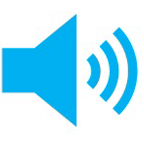}}%
  \endgroup
}

\newcommand{\envChar}{%
  \begingroup\normalfont
  \raisebox{-0.18\height}
  {\includegraphics[height=\symbolHt]{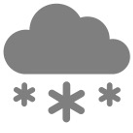}}%
  \endgroup
}
\DeclareMathOperator*{\argmax}{arg\,max}
\def\paperID{*****} %
\def\confName{CVPR}
\def\confYear{2025}
\begin{document}
\title{TaxaBind: A Unified Embedding Space for Ecological Applications}

\author{Srikumar Sastry, Subash Khanal, Aayush Dhakal, Adeel Ahmad, Nathan Jacobs\\
Washington University in St.\ Louis\\
\{{\tt\small s.sastry, k.subash, a.dhakal, aadeel, jacobsn}\}{\tt\small @wustl.edu
}
}
\twocolumn[{%
\renewcommand\twocolumn[1][]{#1}%
\maketitle
\begin{center}
    \centering
    \captionsetup{type=figure}
    \includegraphics[width=\textwidth]{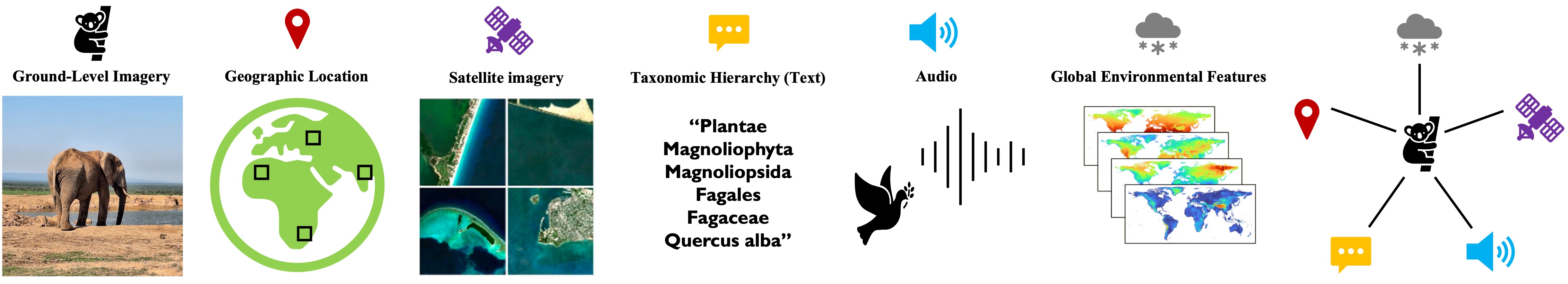}
    \captionof{figure}{\textbf{TaxaBind Framework}. To create a unified embedding space consisting of different modalities, we exploit ground-level images of species as the binding modality. We use various ground-level image-paired datasets to train modality-specific encoders. Ultimately, the encoders support embedding arithmetic and exhibit emergent properties and zero-shot capabilities.}
    \label{img:teaser}
\end{center}%
}]

\maketitle
\begin{abstract}
We present TaxaBind, a unified embedding space for characterizing any species of interest. TaxaBind is a multimodal embedding space across six modalities: ground-level images of species, geographic location, satellite image, text, audio, and environmental features, useful for solving ecological problems. To learn this joint embedding space, we leverage ground-level images of species as a binding modality. %
We propose multimodal patching, a technique for effectively distilling the knowledge from various modalities into the binding modality. We construct two large datasets for pretraining: iSatNat with species images and satellite images, and iSoundNat with species images and audio. Additionally, we introduce TaxaBench-8k, a diverse multimodal dataset with six paired modalities for evaluating deep learning models on ecological tasks. Experiments with TaxaBind demonstrate its strong zero-shot and emergent capabilities on a range of tasks including species classification, cross-model retrieval, and audio classification. The datasets and models are made available at \url{https://github.com/mvrl/TaxaBind}.
\end{abstract}
    
\section{Introduction}
\label{sec:intro}

Fine-grained species classification is a challenging task in computer vision, which is often necessary for ecologists to automatically label images of rare species. A related and arguably a more important task is species distribution mapping which aims to map the presence of a given species of interest. Until now, both tasks were addressed using separate frameworks and methodologies, often requiring different datasets. In this work, we propose learning a unified embedding space over six modalities that is useful for several downstream ecological tasks including but not limited to species distribution mapping, fine-grained classification, and audio classification.

The presence of a particular species at a given geographic location can reveal several important characteristics of that species. Previous studies attempted to implicitly learn the relationship between geographic location and the presence of species by considering either environmental features~\cite{cole2023spatial} or satellite images~\cite{sastry2024birdsat,teng2024satbird,dollinger2024sat} describing the location. This leads to learning an effective representation of any geographic location which is useful for species distribution mapping. However, this type of modeling often overlooks important species attributes, such as their taxonomic hierarchy or audio signatures.

Recent works such as BioCLIP~\cite{stevens2024bioclip} and ArborCLIP~\cite{yang2024arboretum} have demonstrated impressive zero-shot species classification capabilities. However, these frameworks are restricted to image and text modalities, ignoring crucial geographic, audio, and habitat characteristics of species. Multimodal embedding frameworks like ImageBind~\cite{girdhar2023imagebind} and GRAFT~\cite{mall2023remote} have shown that it is possible to learn a joint representation space by aligning all available modalities to the ground-level image modality. This allows for training modality-specific encoders using only image-paired datasets. One potential downside of such methods is that they perform locked tuning with the ground-level image modality. This means that the ground-level image encoder is kept frozen, while the other modalities are trained to project to the existing learned space of the ground-level image modality. This can lead to sub-optimal performance since task-specific unique information of each modality is lost~\cite{liang2024factorized}.

To this end, we propose multimodal patching, building upon patching~\cite{ilharco2022patching}, a framework to distill knowledge from various modalities while still preserving the original embedding space of the binding modality. We show that multimodal patching can improve zero-shot classification performance of the binding modality. We create a joint embedding space containing six modalities~(Figure \ref{img:teaser}).
To facilitate future research and evaluation of ecological models, we present TaxaBench-8k, a truly multimodal dataset containing \textit{six paired modalities}. The contributions of our work are fivefold:
\begin{enumerate}
    \item \textit{Multimodal Patching}. We propose a simple yet effective patching technique that improves over the ImageBind framework.
    \item \textit{Multimodal Models for Ecological Applications}. We propose modality-specific encoders that can handle various ecological tasks over six modalities: ground-level image, geographic location, satellite image, text, audio, and environmental features. 
    \item \textit{Multimodal datasets}. We compiled two large-scale novel cross-view datasets: i) iSoundNat: ground-level images of species with their corresponding audio; and ii) iSatNat: ground-level images of species with their corresponding satellite imagery.
    \item \textit{TaxaBench-8k}. We present TaxaBench-8k, a benchmarking dataset containing six paired modalities for evaluating multimodal ecological models.
    \item We demonstrate our models' effectiveness and emergent properties on several benchmarking and zero-shot tasks.
    
\end{enumerate}
\section{Related Work}
\subsection{Multimodal Self-Supervised Learning}
Multimodal self-supervised methods using contrastive learning (CL) objectives have shown impressive results across diverse tasks. These methods encompass advancements in CL frameworks, integration of multiple modalities into unified embedding spaces, and innovations in training strategies to further enhance model performance. Out of many notable works in this area, CLIP~\cite{radford2021learning} utilizes symmetric InfoNCE loss~\cite{oord2018representation}; SupCon~\cite{khosla2020supervised} utilizes label information in its contrastive objective; SigLIP~\cite{zhai2023sigmoid} employs binary classification loss. Frameworks such as ImageBind~\cite{girdhar2023imagebind}, Sat2Cap~\cite{dhakal2024sat2cap}, GRAFT~\cite{mall2023remote}, GeoCLAP~\cite{khanal2023soundscape}, and GeoBind~\cite{dhakal2024geobind} demonstrate the integration of additional modalities such as audio, satellite imagery, or metadata into CL-trained multimodal embedding spaces. The core idea of these strategies involves utilizing a pretrained image-text embedding space and learning to project all other modalities into this space. However, this simple yet effective strategy can suffer from information collapse~\cite{liang2024factorized}. 

Recent innovations in training strategies include LiT~\cite{zhai2022lit}, which enhances zero-shot performance by freezing the vision encoder while training the text encoder; OmniVec~\cite{srivastava2024omnivec} and OmniVec2~\cite{srivastava2024omnivec2}, which improve performance through modality-specific encoders and shared backbones for multimodal multitask learning; factorized contrastive learning~\cite{liang2024factorized} which focuses on preserving unique modality-specific information; and Patching~\cite{ilharco2022patching}, which boosts performance by interpolating weights between pre-trained and fine-tuned models. In this work, we generalize the concept of patching to more than two modalities. We call this multimodal patching which improves over the training strategy of ImageBind.

\subsection{Multimodal Learning for Ecology}
Multimodal learning for ecological applications, such as species distribution modeling (SDM) and fine-grained visual classification (FGVC) of species, has recently advanced significantly. These advancements are driven by the availability of large-scale multimodal datasets \cite{van2018inaturalist,lorieul2022overview,stevens2024bioclip,yang2024arboretum} and novel multimodal learning frameworks \cite{diao2022metaformer,sastry2024birdsat, sastry2023ld,dollinger2024sat,huynh2024contrastive}. Methods like BioCLIP \cite{stevens2024bioclip} and ArborCLIP \cite{yang2024arboretum} have demonstrated the utility of combining images of species with their corresponding taxonomic text descriptions. BioCLIP is a foundational model for the tree of life, trained with a CLIP-like contrastive loss between image representations of different species and taxonomic descriptions. ArborCLIP is a recent model contrastively trained on a large multimodal dataset containing over 134 million images of diverse species paired with their detailed taxonomic descriptions. Although such CLIP-based models have shown state-of-the-art performance in zero-shot FGVC, they are limited by the number of modalities they can consume and the tasks they can solve. 

On the other hand, the multimodal fusion of remote sensing data has resulted in unprecedented performance in SDM under presence-only \cite{mac2019presence,cole2023spatial,sastry2023ld,dollinger2024sat} and presence-absence \cite{teng2024satbird,picek2024geoplant} settings. Most of these methods are general-purpose ecological predictors that effectively learn multimodal features that can vary across space. In these frameworks, geolocations are represented either by learning an implicit neural function \cite{diao2022metaformer,cole2023spatial,sastry2023ld} or by representations derived from satellite imagery \cite{teng2024satbird,dollinger2024sat,picek2024geoplant}. Other CL-based methods such as SatClip~\cite{klemmer2023satclip} and GeoCLIP~\cite{vivanco2024geoclip} aim to learn general-purpose location representations which can then eventually be utilized for downstream ecological tasks. In our work, we combine these parallel lines of work into a single general-purpose framework, TaxaBind, that can consume multiple modalities and solve numerous ecology-related tasks. 

\section{Dataset}
In this work, we train and evaluate our models using various large-scale multimodal datasets. We built three datasets to advance multimodal learning in the field. Here we provide the details about the datasets.

\textbf{Training datasets}. When it comes to applying deep learning in ecology, there is a lack of high-quality multimodal datasets, especially the ones with paired ground-level images of species. To bridge the gap, we constructed two large-scale datasets: \texttt{iSatNat} and \texttt{iSoundNat} (Table~\ref{tab:datasets}). iSatNat consists of pairs of ground-level and satellite imagery while iSoundNat contains pairs of ground-level images and audio recordings of species. We begin with the iNat-2021 dataset~\cite{van2018inaturalist} which contains 2.7M images of species alongwith metadata including geolocation information. We built iSatNat by collecting \textbf{2.7M} Sentinet-2 level 2A imagery corresponding to each ground-level image of species in the iNat-2021 dataset. iSoundNat was built by collecting ground-level images of species from the iNaturalist platform which contained audio recordings. We specifically downloaded \textit{research grade} observations from the platform. This resulted in a total of \textbf{88,130} pairs of images and audio. For more details about the datasets, please refer to the appendix. We leverage BioCLIP's TreeofLife-10M~\cite{stevens2024bioclip} dataset for image-text pretraining. To simplify our experimentation, we use pretrained BioCLIP vision and text encoders. For training the location encoder, we use the geolocation information present in the iNat-2021 dataset corresponding to each image. We use the environmental variables from WorldClim-2.1 for training the environmental encoder. To do this, we extract the bioclimatic variables corresponding to the geolocations present in the iNat-2021 dataset.

\textbf{TaxaBench-8k}. For evaluation and further research, we constructed a truly multimodal dataset containing six paired modalities: ground-level image, geographic location, satellite image, text,
audio, and environmental features. We begin with the test split of our iSoundNat dataset which contains \textbf{8,813} image and audio pairs. For each sample, we downloaded Sentinel-2 level 2A imagery corresponding to the geolocation information present with that sample. Similarly, we extracted the environmental features from WorldClim-2.1 corresponding to the geolocations of the samples. Our TaxaBench-8k dataset is multifaceted and can be used to evaluate ecological models for various tasks such as cross-modal retrieval, species classification, and audio classification.

\textbf{Evaluation datasets}. We evaluate our models on a range of downstream ecological tasks. We assess the effectiveness of each encoder through these downstream tasks. The details of the datasets are mentioned in Section~\ref{exp} with additional details present in the appendix.

\begin{table}
  \centering
  \begin{center}
  \begin{tabular}{l c c c}
    \toprule
    \textbf{iSoundNat} & Train & Val&Test\\
    \midrule
    \#samples&74,910&4,407&8,813\\
    \#species&6,925&1,482&2,225\\
    \bottomrule
    \toprule
    \textbf{iSatNat} & Train & Val&Test\\
    \midrule
    \#samples&2.55M&134k&100k\\
    \#species&10k&10k&10k\\
    \bottomrule
  \end{tabular}
  \caption{The number of samples and unique species categories in our datasets.}
  \label{tab:datasets}
  \end{center}
\end{table}

\section{Method}
We aim to learn a unified embedding space that can uniquely characterize a given taxon. This is done by aligning all available modalities to the ground-level images of species. For this, we utilize the BioCLIP~\cite{stevens2024bioclip} embedding space as a teacher and learn to project the embeddings from all the modalities to this space. We propose multimodal patching to further distill task-specific unique knowledge into the modalitiy-specific encoders. This enables embedding arithmetic which is useful for zero-shot classification and cross-modal retrieval tasks. Further, the modalities exhibit emergent properties with each other when only trained with corresponding ground-level image pairs. Below we discuss the contrastive learning approach used for training, multimodal patching, and the implementation details of our framework.

\subsection{Contrastive Learning}
InfoNCE~\cite{oord2018representation} is a widely used loss function for learning an embedding space with similar and dissimilar examples within a single mini-batch. This loss function works by treating paired examples as positives while considering all other pairs as negatives. %
These pairs can be constructed using augmentation techniques or can originate from different modalities, such as images and text. However, for our problem, we note that a single mini-batch may contain multiple instances of the same species category. Considering examples from the same species category as negative may lead to sub-optimal performance. Hence, we utilize the concept of CLIP loss~\cite{radford2021learning} and combine it with SupCon loss~\cite{khosla2020supervised} as the species category information is available for each ground-level image. Consider the pair of modalities ($\mathcal{G}$, $\mathcal{M}$), where $\mathcal{G}$ denotes the ground-level image modality while $\mathcal{M}$ denotes some other modality (e.g. satellite imagery). Consider a dataset with aligned observations of the two modalities and a mini-batch of examples during training $\{g_i,m_i\}_{i=1,\ldots N}$. Using deep networks $f$ and $h$, one can obtain normalized embeddings of the respective modalities as $z_i=f(g_i)$ and $y_i=h(m_i)$. Let $N(i)$ denote the set of all plausible indices in a mini-batch and $P(i)$ denote the set of indices of examples with the same species category as that of the $i^{th}$ example. Then, the deep networks are optimized using the following combined loss:
\begin{align}
    \mathcal{L}_{\mathcal{G}\xrightarrow{}\mathcal{M}} = \sum_{i\in N(i)}{\frac{-1}{|P(i)|}\sum_{j\in P(i)}{\text{log}\frac{\text{exp}(z_i \boldsymbol{\cdot} y_j/\tau)}{\sum_{n\in N(i)}{\text{exp}(z_i \boldsymbol{\cdot} y_n/\tau)}}}}
    \nonumber\\
    \mathcal{L}_{\mathcal{M}\xrightarrow{}\mathcal{G}} = \sum_{i\in N(i)}{\frac{-1}{|P(i)|}\sum_{j\in P(i)}{\text{log}\frac{\text{exp}(y_i \boldsymbol{\cdot} z_j/\tau)}{\sum_{n\in N(i)}{\text{exp}(y_i \boldsymbol{\cdot} z_n/\tau)}}}}
    \nonumber
\end{align}
\begin{equation}
    \mathcal{L}_{\mathcal{G},\mathcal{M}} = \frac{\mathcal{L}_{\mathcal{G}\xrightarrow{}\mathcal{M}} + \mathcal{L}_{\mathcal{M}\xrightarrow{}\mathcal{G}}}{2}
\end{equation}
where $\tau$ is a scalar temperature parameter that controls the sensitivity of the predicted softmax distribution. The loss function seeks to cluster positive examples from the modalities in the embedding space while pushing negative examples farther away.

\begin{figure}[t]
  \centering
  \includegraphics[width=\linewidth]{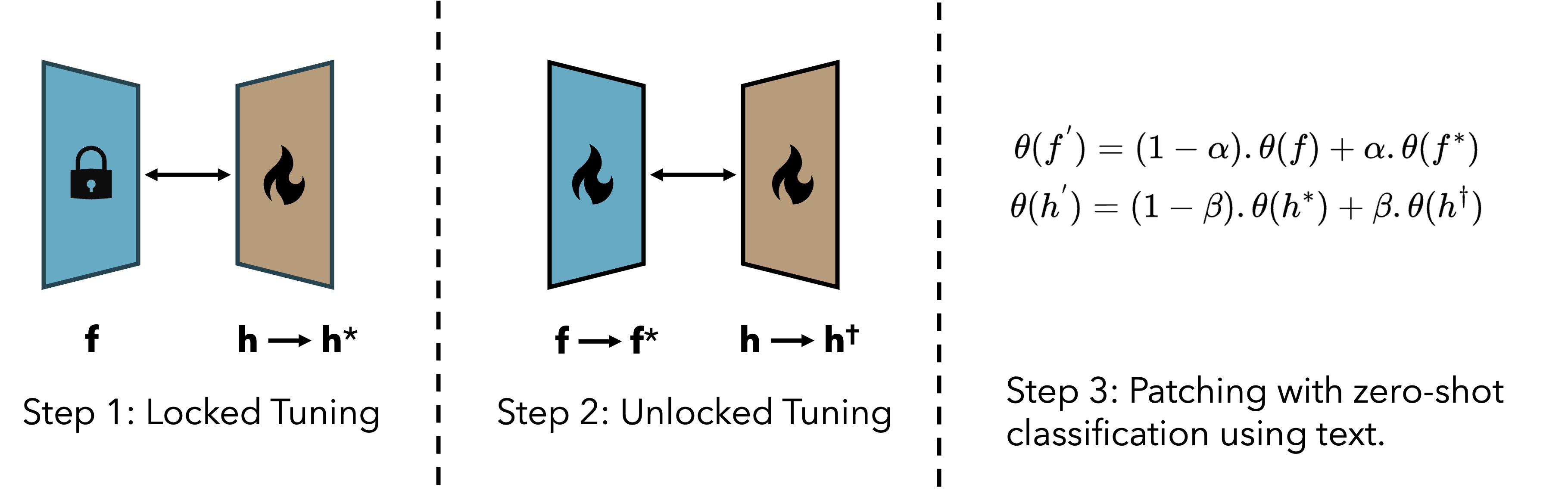}
   \caption{\textbf{Multimodal Patching}. For distilling unique information from different modalities, we patch the encoders using zero-shot classification with text. Note that since the network $f$ is shared across all modalities, it is patched using techniques like sequential patching or parallel patching. %
}
   \label{fig:patching}
\end{figure}

\subsection{Multimodal Patching}
We describe multimodal patching as shown in Figure~\ref{fig:patching}. The overall framework consists of three steps. The pseudo-code for multimodal patching is presented in Algorithm~\ref{alg:mmp}. Let $f$ be a pretrained binding modality encoder and $h$ be an encoder of a different modality. First, we perform locked tuning of $h$ to obtain $h^*$. In this step, $f$ is utilized as a teacher, and embeddings obtained from $h$ are learned to project to the existing embedding space of $f$. In the second step, we perform full finetuning of $f$ and $h$ to obtain $f^*$ and $h^\dagger$ respectively. The encoders $f^*$ and $h^\dagger$ no longer preserve the original embedding space of $f$. Finally, in the third step, we perform patching~\cite{ilharco2022patching} of $h$ by linearly interpolating the weights between $h^*$ and $h^\dagger$. The interpolation weights are determined by analyzing the performance of the patched models on a patching task $\mathcal{P}$. We utilize zero-shot classification with text as the patching task. This patching task helps to preserve the original embedding space of $f$ and enables emergent capabilities. These steps are repeated for all available modalities. Since $f$ is shared across the modalities, we perform sequential patching of $f$ across all the modalities as described in Algorithm~\ref{alg:mmp}. Our strategy modifies the weights of $f$ while preserving the original embedding space, unlike ImageBind, where $f$ is frozen. 

\begin{algorithm}[t]
\caption{Multimodal Patching}\label{alg:mmp}
\begin{algorithmic}[1]
\REQUIRE{Binding Modality $\mathcal{G}$, Set of Modalities $\mathcal{M}$, Binding Modality Encoder $f$, Set of Encoders of $\mathcal{M}$ - $\{h_i\}_{i=1,\ldots|\mathcal{M}|}$}, Patching Task $\mathcal{P}$
\vspace{1mm}
\STATE $\theta(z) \leftarrow \theta(f)$
\FOR{\textbf{each} $h_i$ in $\{h_i\}_{i=1,\ldots|\mathcal{M}|}$}
\STATE $h_i^* \leftarrow \text{locked\_tuning(}f,h_i\text{)}$
\STATE $f^*,h_i^\dagger \leftarrow \text{unlocked\_tuning(}f,h_i\text{)}$\\
\STATE $\alpha \leftarrow \argmax_{\alpha} \mathcal{P}[(1 - \alpha).\theta(z) + \alpha.\theta(f^*)]$\\
\STATE $\beta \leftarrow \argmax_{\beta} \mathcal{P}[(1 - \beta).\theta(h_i^*) + \beta.\theta(h_i^\dagger)]$\\
\STATE $\theta(z) \leftarrow (1 - \alpha).\theta(z) + \alpha.\theta(f^*)$
\STATE $\theta(h_i^{'}) \leftarrow (1 - \beta).\theta(h_i^*) + \beta.\theta(h_i^\dagger)$
\ENDFOR
\STATE $\theta(f^{'}) \leftarrow \theta(z)$
\RETURN $f^{'}$, $\{h_i^{'}\}_{i=1,\ldots|\mathcal{M}|}$
\end{algorithmic}
\end{algorithm}

\subsection{Implementation Details}
\label{impdet}
We aim to learn an embedding space consisting of six modalities. To do this, we employ modality-specific encoders to encode each of the modalities into a joint embedding space. This setup proves to be computationally feasible and allows for precomputing embeddings. We use the transformer architecture for encoding ground-level images, text, satellite images, and audio. We use pretrained BioCLIP~\cite{stevens2024bioclip} (ViT-B/16) vision and text encoders for the ground-level images and text respectively. We use pretrained CLIP~\cite{radford2021learning} (ViT-B/16) vision encoder for the satellite images and a pretrained CLAP~\cite{elizalde2023clap} encoder for audio. We use the architecture described in GeoCLIP~\cite{vivanco2024geoclip} to encode the geographic location. It consists of an Equal Earth Projection (EEP) operation, a random Fourier feature transform, and an MLP network. We use a ResNet-style MLP~\cite{cole2023spatial} to encode the environment features for a given geographic location. 

\begin{figure*}[!t]
\centering
\begin{subfigure}{.24\linewidth}
  \centering
  \includegraphics[width=\columnwidth]{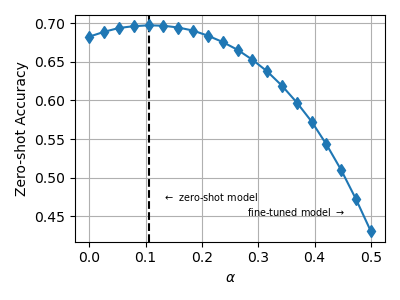}
  \caption{\imageChar\contChar\locChar}
  \label{fig:sub1}
\end{subfigure}%
\begin{subfigure}{.24\linewidth}
  \centering
  \includegraphics[width=\columnwidth]{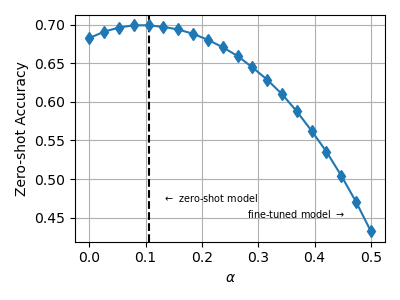}
  \caption{\imageChar\contChar\satChar}
  \label{fig:sub2}
\end{subfigure}
\begin{subfigure}{.24\linewidth}
  \centering
  \includegraphics[width=\columnwidth]{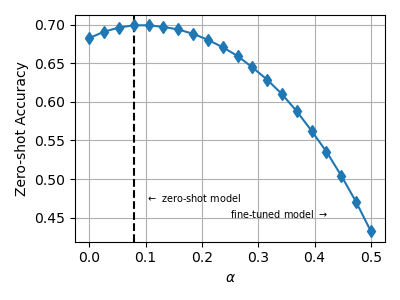}
  \caption{\imageChar\contChar\envChar}
  \label{fig:sub3}
\end{subfigure}
\begin{subfigure}{.24\linewidth}
  \centering
  \includegraphics[width=\columnwidth]{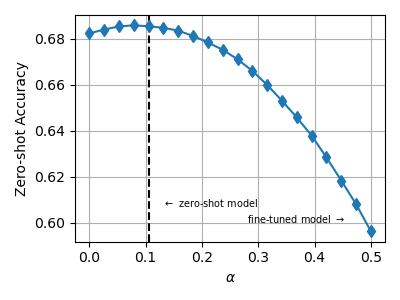}
  \caption{\imageChar\contChar\audioChar}
  \label{fig:sub4}
\end{subfigure}
\caption{\textbf{Patching improves zero-shot classification performance with text}. We evaluate the zero-shot classification accuracy of the ground-level image encoder with different values of $\alpha$ on iNat-2021. We observe performance improvements in all the cases.}
\label{fig:zspatching}
\end{figure*}
\begin{table*}[!ht]
  \centering
  \begin{center}
  \begin{tabular}{lcccccc}
    \toprule
    Model & Modality &Birds525& CUB-200-2011& BioCLIP-Rare& iNat-2021&TaxaBench-8k\\
    \midrule
    BioCLIP~\cite{stevens2024bioclip}&\imageChar&82.92&77.51&34.52&68.24&32.88\\
   ArborCLIP~\cite{yang2024arboretum}&\imageChar&65.84&\textbf{82.41}&27.58&68.00&31.34\\
   \midrule
   TaxaBind&\imageChar&\textbf{83.74}&78.22&\textbf{35.84}&\textbf{70.09}&\textbf{34.45}\\
    \midrule
    \multirow{4}{*}{ImageBind~\cite{girdhar2023imagebind}} &\imageChar\plusChar\locChar&-&-&-&71.02&36.40\\
    &\imageChar\plusChar\satChar&-&-&-&72.62&36.30\\
    &\imageChar\plusChar\envChar&-&-&-&71.96&\textbf{36.59}\\
    &\imageChar\plusChar\audioChar&-&-&-&-&35.91\\
    \midrule
    \multirow{4}{*}{TaxaBind}&\imageChar\plusChar\locChar&-&-&-&\textbf{72.73}&\textbf{36.59}\\
    &\imageChar\plusChar\satChar&-&-&-&\textbf{73.20}&\textbf{37.54}\\
    &\imageChar\plusChar\envChar&-&-&-&\textbf{72.02}&36.51\\
    &\imageChar\plusChar\audioChar&-&-&-&-&\textbf{36.27}\\
    \bottomrule
  \end{tabular}
  \caption{Zero-shot classification performance on various fine-grained species classification datasets using the taxonomic description of species.}
  \label{tab:zero-shot}
  \end{center}
\end{table*}

We use a native embedding size of 512 for all the encoders as used in BioCLIP. All the encoders are trained independently using the BioCLIP vision encoder. Images are normalized and resized to (224, 224) pixels before feeding them to their respective encoders. For training the location encoder, we additionally sample pseudo-negative locations as a way to train on locations absent in the training dataset~\cite{cole2023spatial}. For zero-shot classification using text, we consider the entire taxonomic description of a species. 
For each audio sample, we average across the channel dimension to get single-channel audio. Then we convert each single channel audio sample into mel-spectrogram features using the default CLAP settings:~\texttt{\{feature\_size=64, sampling\_rate=48000, hop\_length=480, max\_length\_s=10, fft\_window\_size=1024\}} in the \texttt{HuggingFace}-wrapper: \texttt{ClapProcessor} for the pre-trained CLAP model~\texttt{clap-htsat-fused}. For each training run, we use 2 NVIDIA H100 GPUs, a batch size of 256, and a gradient accumulation of 8.

\section{Experiments}
\label{exp}
We conduct several empirical evaluations of our modality-specific encoders against state-of-the-art methods across a range of ecology-related tasks. All details about each dataset used for evaluation are mentioned in appendix. Additionally, to evaluate the effectiveness of our proposed multimodal patching strategy, we compare it against the training recipe of ImageBind~\cite{girdhar2023imagebind}, where the binding modality encoder is kept frozen during the training. The ImageBind training recipe is equivalent to restricting our multimodal patching strategy to its first step. In the experiments, we show that adding the embeddings from multiple modalities leads to improved performance compared to using embeddings of a single modality. Below we discuss various experimental results.

\begin{figure*}[t]
  \centering
  \includegraphics[width=\linewidth]{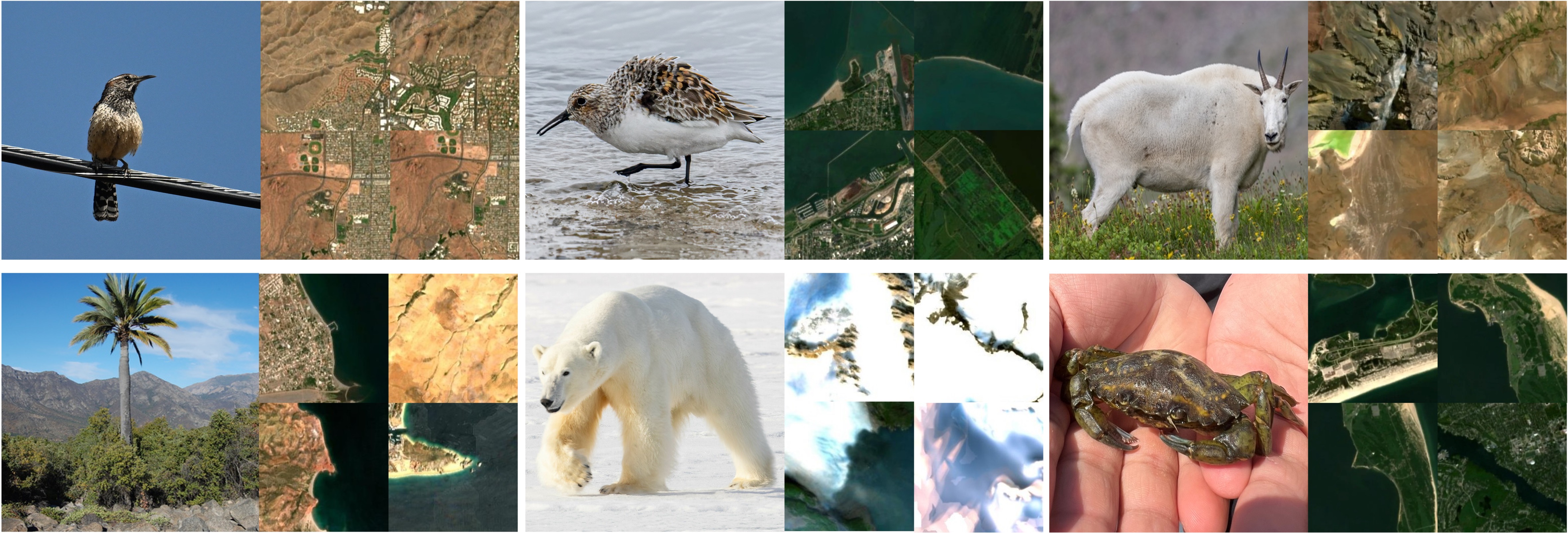}
   \caption{\textbf{Species image to satellite image retrieval task}. For each example, we show the top 4 most similar satellite images retrieved by our model from a gallery of 100k satellite images in the iSatNat-test set.}
   \label{fig:sp2sat}
\end{figure*}

\textbf{Multimodal Patching}. We first show that patching improves zero-shot image classification of the binding modality encoder. In Figure~\ref{fig:zspatching}, we present the zero-shot image classification performance of our model when patched with varying values of $\alpha$. It is seen that for certain values of $\alpha$, our model exhibits improved zero-shot performance than the base pretrained model of ImageBind. This behavior is observed across all the modalities independently. For all the modalities, the optimal value of $\alpha$ is close to 0.1, indicating that models patched using higher values of $\alpha$ (close to 1) move significantly away from the embedding space of BioCLIP. We further show in Table~\ref{tab:singlepatch} that multimodal sequential patching has an additive effect on our model. These experiments empirically demonstrate that ImageBind's training method fails to effectively distill knowledge from various modalities and that it is possible to improve the performance of pretrained binding modality encoder.

\textbf{Zero-shot image classification}. We evaluate TaxaBind's image encoder on zero-shot image classification task using the taxonomic description of species. We use BioCLIP~\cite{stevens2024bioclip} and ArborCLIP~\cite{yang2024arboretum} as image-only baseline models. For multimodal evaluation, we compare against our ImageBind-trained baseline. In this setting, the embeddings obtained from the image encoder are added to those obtained from a different modality encoder and then used for zero-shot classification.
For image-only experiments, we use Birds525~\cite{Piosenka_2023}, CUB-200-2011~\cite{reed2016learning} and BioCLIP-Rare~\cite{stevens2024bioclip}. For multimodal experiments, we use iNat-2021~\cite{van2018inaturalist} and TaxaBench-8k. The results presented in Table~\ref{tab:zero-shot} show that our model outperforms the baseline models in both settings across 4 out of 5 datasets. It is worth noting that our model always outperforms BioCLIP, which indicates the benefit of distilling multimodal information into the ground-level image encoder. Finally, the addition of ground-level and satellite image embedding leads to the best zero-shot classification performance.

\begin{table}[!ht]
  \begin{center}
  \resizebox{\columnwidth}{!}{%
  \begin{tabularx}{1.001\columnwidth}{lcccc}
    \toprule
    Method & Modality & R@1 & R@5 & R@10\\
    \midrule
    \textit{Random Baseline}&-&0.01&0.05&0.11\\
    \midrule
     \multirow{4}{*}{ImageBind~\cite{girdhar2023imagebind}}&\satChar{}\arrowChar{}\locChar&\textbf{8.79}&\textbf{22.72}&\textbf{30.84}\\
     &\locChar{}\arrowChar{}\satChar{}&9.32&24.16&32.24\\
     &\satChar{}\arrowChar{}\audioChar{}&1.94&6.68&10.56\\
     &\audioChar{}\arrowChar{}\satChar{}&1.86&5.33&9.05\\
     \midrule
     \multirow{4}{*}{TaxaBind}&\satChar{}\arrowChar{}\locChar&8.43&21.72&30.42\\
    &\locChar{}\arrowChar{}\satChar{}&\textbf{9.62}&\textbf{24.60}&\textbf{33.42}\\
     &\satChar{}\arrowChar{}\audioChar{}&\textbf{2.05}&\textbf{7.03}&\textbf{11.05}\\
     &\audioChar{}\arrowChar{}\satChar{}&\textbf{2.36}&\textbf{5.96}&\textbf{9.50}\\
    \bottomrule
  \end{tabularx}}
  \caption{\textbf{Emergent capabilities}. Cross-modal retrieval results on the TaxaBench-8k dataset. The gallery consists of 8,813 samples in each experiment.}
  \label{tab:retr}
  \end{center}
\end{table}

\begin{table*}[!ht]
  \centering
  \begin{tabular}{lcccc}
    \toprule
    Model & Modality & BirdCLEF-2022 (\%)&BirdCLEF-2023 (\%)&BirdCLEF-2024 (\%)\\
    \midrule
    CLAP~\cite{elizalde2023clap}&\audioChar&42.33&32.85&39.72\\
    \midrule
    ImageBind&\audioChar&47.11&37.46&45.04\\
    TaxaBind&\audioChar&\textbf{52.60}&\textbf{42.19}&\textbf{49.31}\\
    \midrule
    ImageBind&\audioChar\plusChar\locChar&60.22&44.04&51.64\\
    TaxaBind&\audioChar\plusChar\locChar&\textbf{65.07}&\textbf{46.97}&\textbf{56.24}\\
    \bottomrule
  \end{tabular}
  \caption{Top-1 linear probing results on the task of bird species audio classification.}
  \label{tab:audiocls}
\end{table*}

\begin{table*}[!ht]
  \centering
  \begin{tabular}{lccccc}
    \toprule
    Model &Modality& Geo Feature (R2) & Ecoregions (\%)&Biomes (\%) & GeoPlant (MSE)\\
    \midrule
    GeoCLIP~\cite{vivanco2024geoclip}&\locChar&75.07&72.18&69.69&0.0472\\
    SatClip~\cite{klemmer2023satclip}&\locChar&74.91&70.08&68.54&0.0594\\
    SINR~\cite{cole2023spatial} &\locChar&\textbf{75.90}&68.04&70.22&0.0467\\
    \midrule
    ImageBind&\locChar&74.51&73.74&\textbf{71.73}&0.0449\\
    TaxaBind&\locChar&74.55&\textbf{73.75}&\textbf{71.73}&\textbf{0.0420}\\
    \bottomrule
  \end{tabular}
  \caption{Linear probing results on various geo-aware ecological tasks to demonstrate the effectiveness of our location encoder.}
  \label{tab:loccls}
\end{table*}

\begin{table*}[!ht]
  \centering
  \begin{tabular}{lcccc}
    \toprule
    Model & Modality & SatBird-Kenya (MSE) & SatBird-USA-sum (MSE) & SatBird-USA-win (MSE) \\
    \midrule
    CLIP~\cite{radford2021learning}&\satChar&0.0832&0.0715&0.0755\\
    RVSA~\cite{wang2022advancing}&\satChar&0.0842&0.0742&0.0791\\
    ScaleMAE~\cite{reed2023scale}&\satChar&0.1200&0.1011&0.0994\\
    SatMAE++~\cite{noman2024rethinking}&\satChar&0.0953&0.0854&0.0867\\
    Sat2Cap~\cite{dhakal2024sat2cap}&\satChar&0.0836&0.0734&0.0770\\
    \midrule
    Imagebind&\satChar&0.0753&0.0662&0.0681\\
    TaxaBind&\satChar&\textbf{0.0721}&\textbf{0.0632}&\textbf{0.0661}\\
    \midrule
    ImageBind&\satChar\plusChar\locChar&0.0730&0.0648&0.0669\\

    TaxaBind&\satChar\plusChar\locChar&\textbf{0.0710}&\textbf{0.0614}&\textbf{0.0642}\\
    \bottomrule
  \end{tabular}
  \caption{Linear probing results on species encounter rates prediction using satellite imagery as input on the SatBird dataset.}
  \label{tab:satbird}
\end{table*}

\textbf{Cross-modal retrieval}. To demonstrate the emergent capabilities of our models, we evaluate on the task of cross-modal retrieval. We use TaxaBench-8k, which has a gallery size of 8,813 data points. In each retrieval setting, we selected the pair of modalities which were not explicitly trained together. The results, shown in Table~\ref{tab:retr}, highlight the superior performance of TaxaBind compared to both a random baseline and the ImageBind framework. We further present six qualitative examples of species image to satellite image retrieval results in Figure~\ref{fig:sp2sat}. Here, we use a gallery of 100k satellite images and retrieve the top 4 most similar satellite images given a ground-level species image. The retrieved images show habitat characteristics correlated with those of the query species. This suggests that our models have learned to capture fine-grained habitat and climate-related information about the species. We present additional quantitative retrieval results in the appendix.

\textbf{Species audio classification}. We assess the effectiveness of our audio encoder on the task of bird species audio classification. We used pretrained CLAP~\cite{elizalde2023clap} and ImageBind's audio encoder as baselines and tested them on three datasets: BirdCLEF-2022, BirdCLEF-2023, and BirdCLEF-2024. These datasets are part of the LifeCLEF series~\cite{joly2022overview,joly2023overview}, which aims to identify bird species based on their soundscape. For each dataset, we perform linear probing over obtained audio embeddings from each audio encoder. The same audio preprocessing pipeline is used as described in Section~\ref{impdet}. We report the top-1 accuracy of the models in Table~\ref{tab:audiocls}. The results demonstrate the effectiveness of our audio encoder in representing the soundscape of species and characterizing the fine-grained differences between them. In the multimodal setting, we add the embeddings of audio and geographic location and then perform linear probing. It is observed that combining the embedding results in an improved performance.

\textbf{Effectiveness of location encoder}. We assess the capability of our location encoder to reason about the ecological traits of a given geographic location. This is important in several downstream applications such as species distribution modeling or climate prediction. To do this, we consider four tasks: 1) geo-feature regression~\cite{cole2023spatial}; 2) ecoregion classification~\cite{dinerstein2017ecoregion}; 3) biome classification~\cite{dinerstein2017ecoregion}; and 4) presence-absence prediction of plant species~\cite{picek2024geoplant}. We use GeoCLIP~\cite{vivanco2024geoclip}, SatCLIP~\cite{klemmer2023satclip} and SINR~\cite{cole2023spatial} as the baseline location encoders. Results reported in Table~\ref{tab:loccls} show that our model outperforms the baseline models in ecoregion, biome, and plant species prediction tasks while also achieving competitive results on the geo-feature regression task. This highlights the effectiveness of our location encoder in addressing a range of geospatial ecological tasks based solely on geographic location input.

\begin{figure*}[!t]
\centering
\begin{subfigure}{.33\linewidth}
  \centering
  \includegraphics[width=\columnwidth]{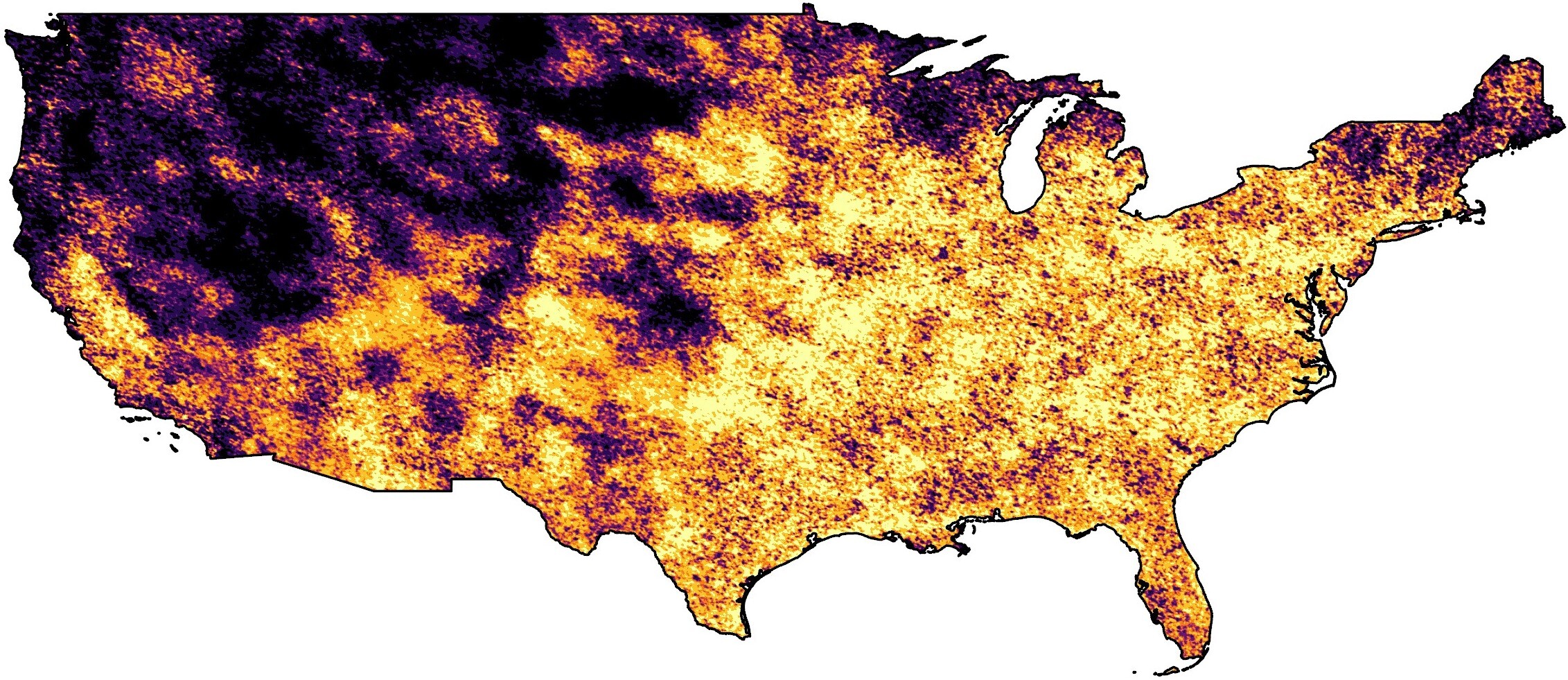}
  \caption{\imageChar\arrowChar\locChar}
\end{subfigure}%
\begin{subfigure}{.33\linewidth}
  \centering
  \includegraphics[width=\columnwidth]{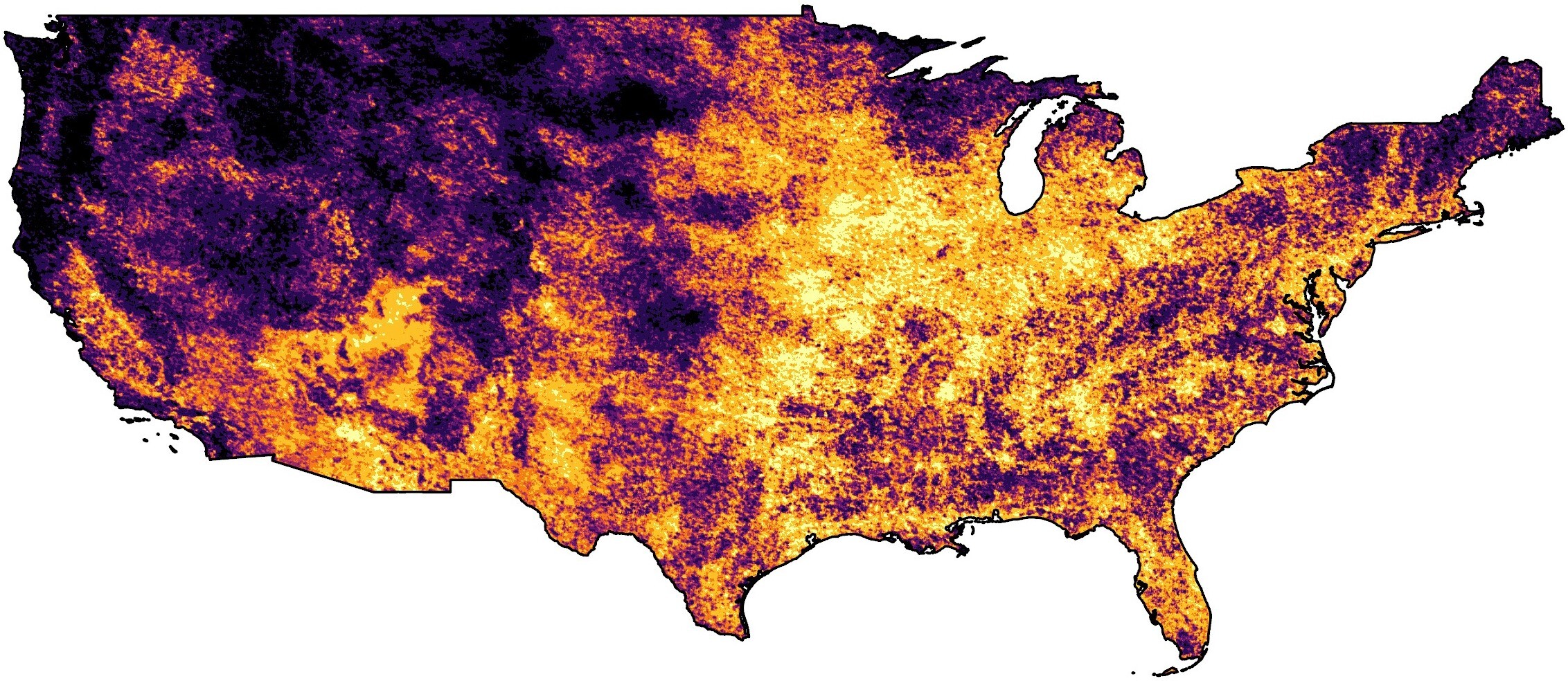}
  \caption{\imageChar\arrowChar\locChar\plusChar\satChar}
\end{subfigure}
\begin{subfigure}{.33\linewidth}
  \centering
  \includegraphics[width=\columnwidth]{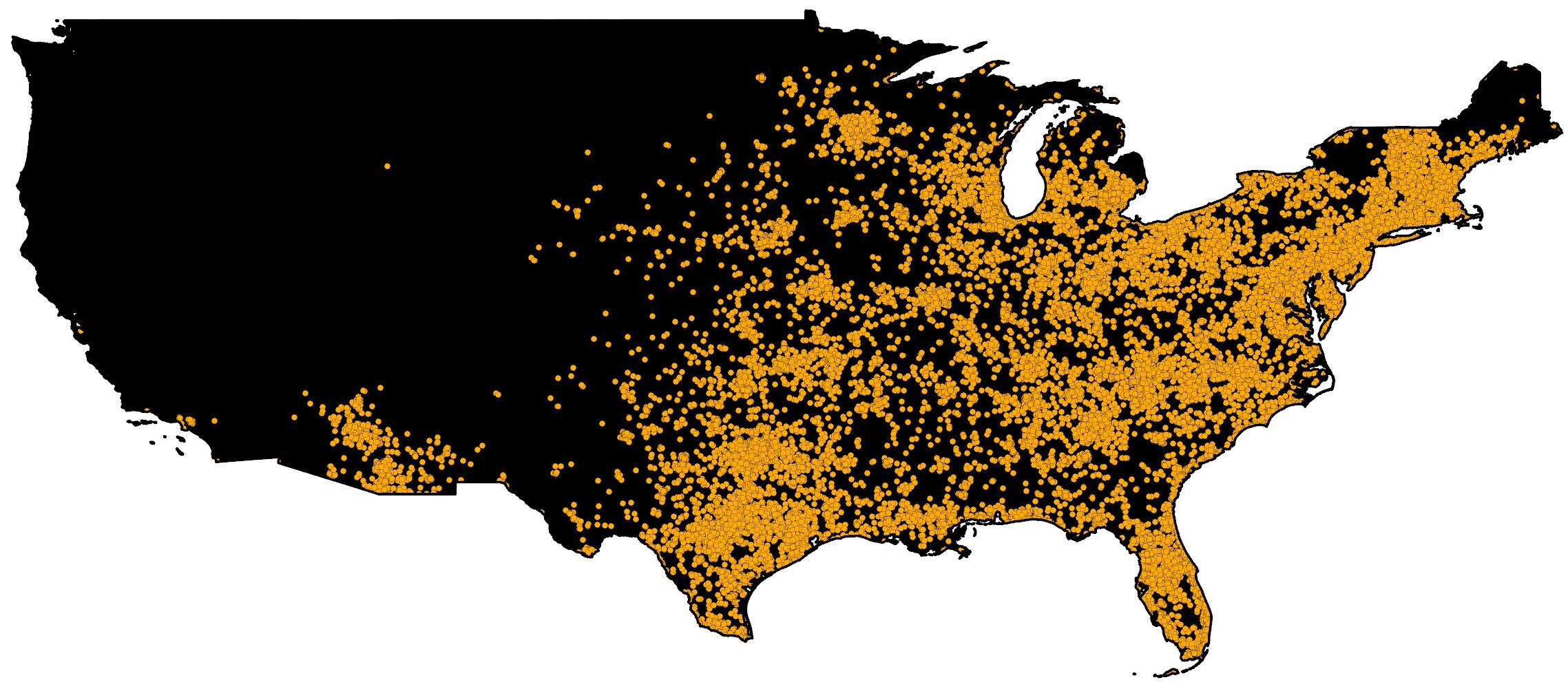}
  \caption{GBIF presence-only}
\end{subfigure}
\caption{\textbf{Zero-shot Species Distribution Map}. We create a species distribution map of \textit{Cardinalis cardinalis} using a query ground-level image and combination of various modalities across the USA.}
\label{fig:sdm}
\end{figure*}

\textbf{Effectiveness of satellite image encoder}. Satellite imagery can provide important visual characteristics of a location which can be used for addressing fine-grained ecological problems. We evaluate our satellite image encoder to predict bird species encounter rates. We employ the SatBird dataset~\cite{teng2024satbird} and follow their training procedure. We perform linear probing on the embeddings obtained from the satellite image encoder. We compare against several state-of-the-art satellite image encoders and report the results in Table~\ref{tab:satbird}. TaxaBind outperforms all the baselines considered for this task. This empirical evidence demonstrates that our model has effectively learned to extract fine-grained ecological traits from satellite imagery.

\textbf{Species range mapping}. We utilize our models to create fine-grained species distribution maps. This is done by computing the similarity between the embeddings of a query ground-level image of species and the embeddings obtained from various modalities for a particular region of interest. The benefit of this approach is that the embeddings for a region can be pre-computed and stored for real-time applications. In Figure~\ref{fig:sdm}, we show species distribution maps generated for \textit{Cardinalis cardinalis} over the USA. We downloaded satellite imagery using a dense grid draped over the USA. Qualitatively, we observe that the generated maps are accurate and that using satellite imagery helps to create more fine-grained maps. 

\subsection{Ablation Studies}
The results reported previously demonstrated that our proposed multimodal patching strategy outperforms the ImageBind training strategy. We now study the performance of using different patching strategies below.

\textbf{Single-modality patching}. In Figure~\ref{fig:zspatching}, we presented the results achieved by the binding modality encoder when patched by training with each modality independently. Now, we investigate whether our sequentially patched model can outperform these single-modality-specific patched models. In Table~\ref{tab:singlepatch}, it is noticed that our model outperforms each of the other patched models. The sequential patching method has an \textit{additive effect} and can distill knowledge from various modalities.

\textbf{Multimodal patching type}. In addition to the sequential patching technique, the parallel patching technique can be used to simultaneously patch the binding modality encoder with all the modalities. This involves averaging the weights of all the single-modality patched models and then determining the best interpolation weight using the patching task. Table~\ref{tab:singlepatch} demonstrates that parallel patching yields inferior results compared to the sequential patching method, and it even performs worse than models patched using geographic location or satellite images. This indicates that averaging the weights of the individually patched models is not optimal. It also suggests that modalities such as geographic location and satellite imagery are more crucial for the downstream zero-shot classification task.

\begin{table}[!ht]
  \begin{center}
  \begin{tabular}{lcccc}
    \toprule
    Method &iNat-2021 & TaxaBench-8k\\
    \midrule
     \imageChar\contChar\locChar&69.70&33.31\\
     \imageChar\contChar\satChar&69.93&34.04\\
     \imageChar\contChar\envChar&68.84&33.30\\
     \imageChar\contChar\audioChar&68.61&33.44\\
     \midrule
     \textit{parallel patching}&69.66&33.64\\
     \textit{sequential patching}&\textbf{70.09}&\textbf{34.45}\\
    \bottomrule
  \end{tabular}
  \caption{The Table shows the additive effect of sequential patching, improving upon single-modality patching.}
  \label{tab:singlepatch}
  \end{center}
\end{table}

\section{Conclusions}
In this work, we presented TaxaBind, a unified embedding space for ecological applications covering six modalities: ground-level images, geographic location, satellite images, text, audio, and environmental features. We introduced multimodal patching, a technique to distill knowledge from multiple modalities into a binding modality, improving upon existing methods like ImageBind. We curated two datasets, iSatNat and iSoundNat, to train our models, and introduced TaxaBench-8k, a multimodal dataset for evaluating ecological models. Our extensive experiments demonstrated TaxaBind's effectiveness on various tasks such as zero-shot species classification, cross-modal retrieval, and audio classification, outperforming state-of-the-art methods. Through our multimodal framework, we showed the effectiveness of combining multiple modalities for addressing downstream ecological tasks. We emphasize that our models are general purpose and could potentially be used for other ecology and climate-related applications such as deforestation mapping, tree canopy height mapping, etc. In the future, we plan to explore different techniques for integrating multimodal data in ecology.

\section{Acknowledgements}
We gratefully acknowledge the Taylor Geospatial Institute and the University of Illinois for providing access to TGI RAILS, a high-performance computing system supported by the National Science Foundation under Grant No. OAC-2232860.

{
    \small
    \bibliographystyle{ieeetr}
    \bibliography{main}
}

\clearpage
\setcounter{page}{1}

\thispagestyle{empty}
\twocolumn[{%
\renewcommand\twocolumn[1][]{#1}%
\maketitlesupplementary
\begin{center}
    \centering
    \captionsetup{type=figure}
    \includegraphics[width=\textwidth]{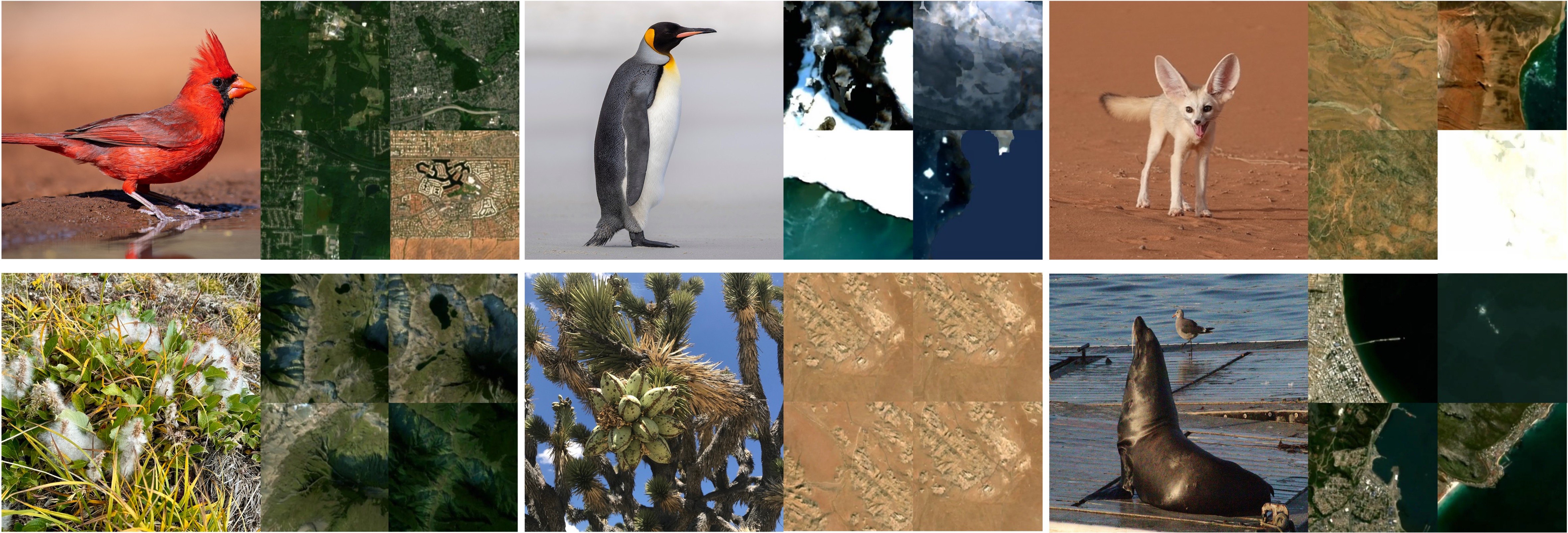}
    \captionof{figure}{Additional results for species image to satellite image retrieval task. For each example, we show the top 4 most similar satellite images retrieved by our model from a gallery of 100k satellite images in the iSatNat-test set.}
    \label{fig:sp2sat_v2}
\end{center}%
}]

\appendix
\section{Retrieval Results}
\begin{figure*}[!t]
\centering
\begin{subfigure}{.33\linewidth}
  \centering
  \includegraphics[width=\columnwidth]{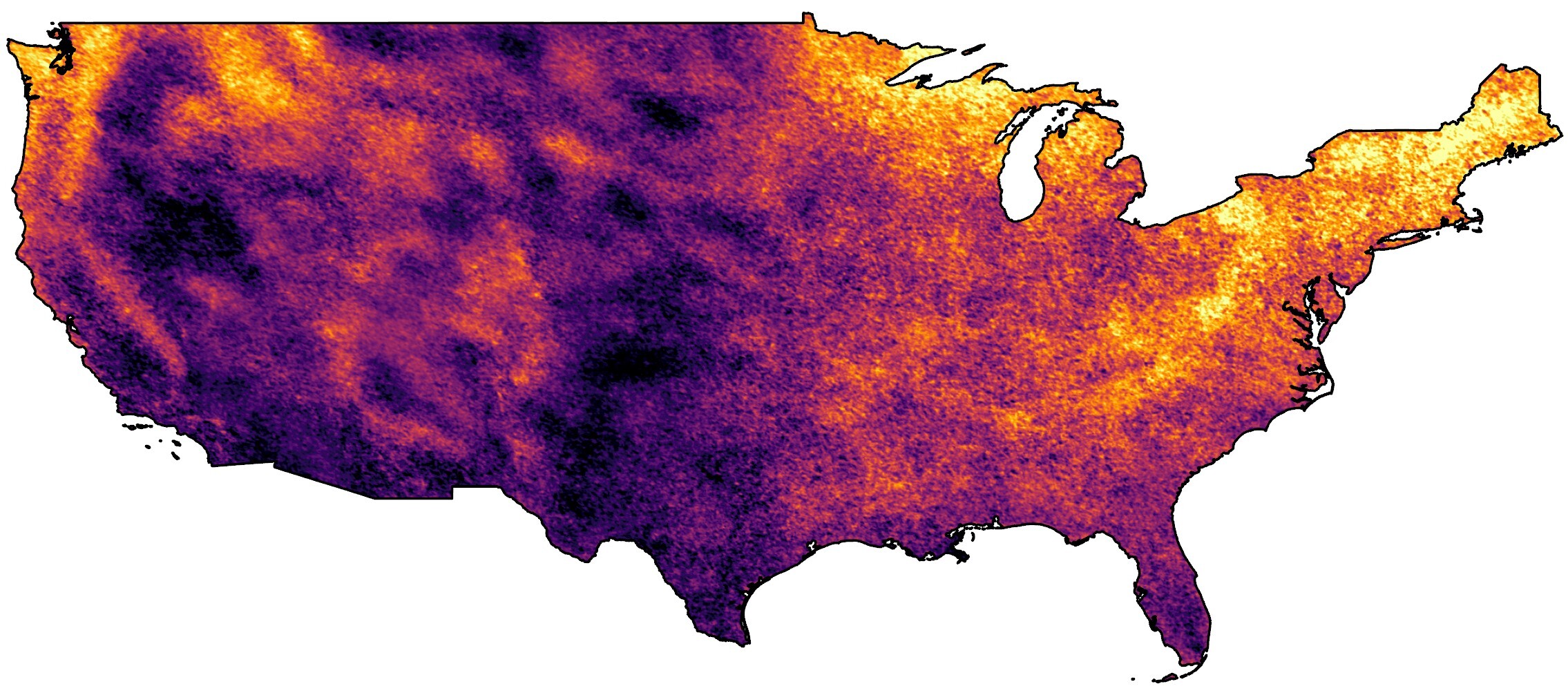}
  \caption{\imageChar\arrowChar\locChar}
\end{subfigure}%
\begin{subfigure}{.33\linewidth}
  \centering
  \includegraphics[width=\columnwidth]{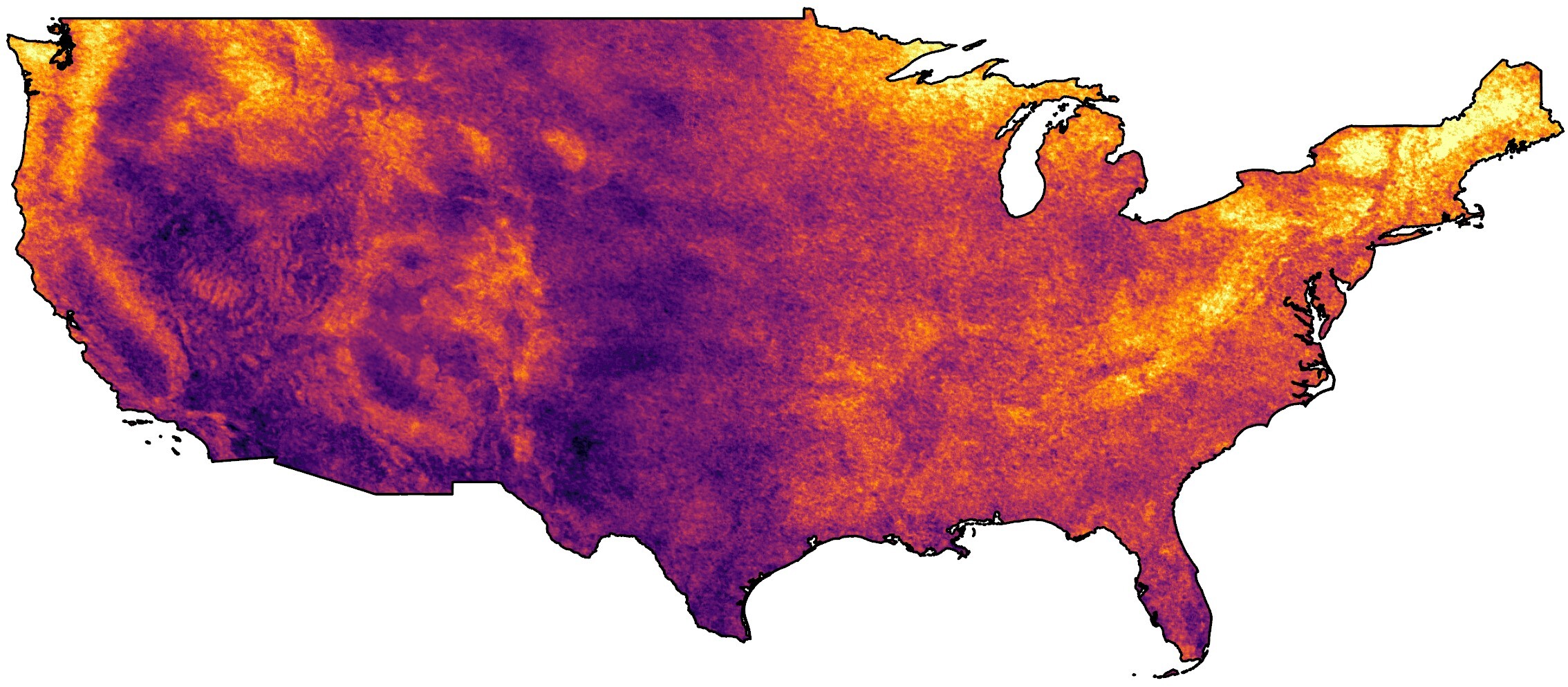}
  \caption{\imageChar\arrowChar\locChar\plusChar\satChar}
\end{subfigure}
\begin{subfigure}{.33\linewidth}
  \centering
  \includegraphics[width=\columnwidth]{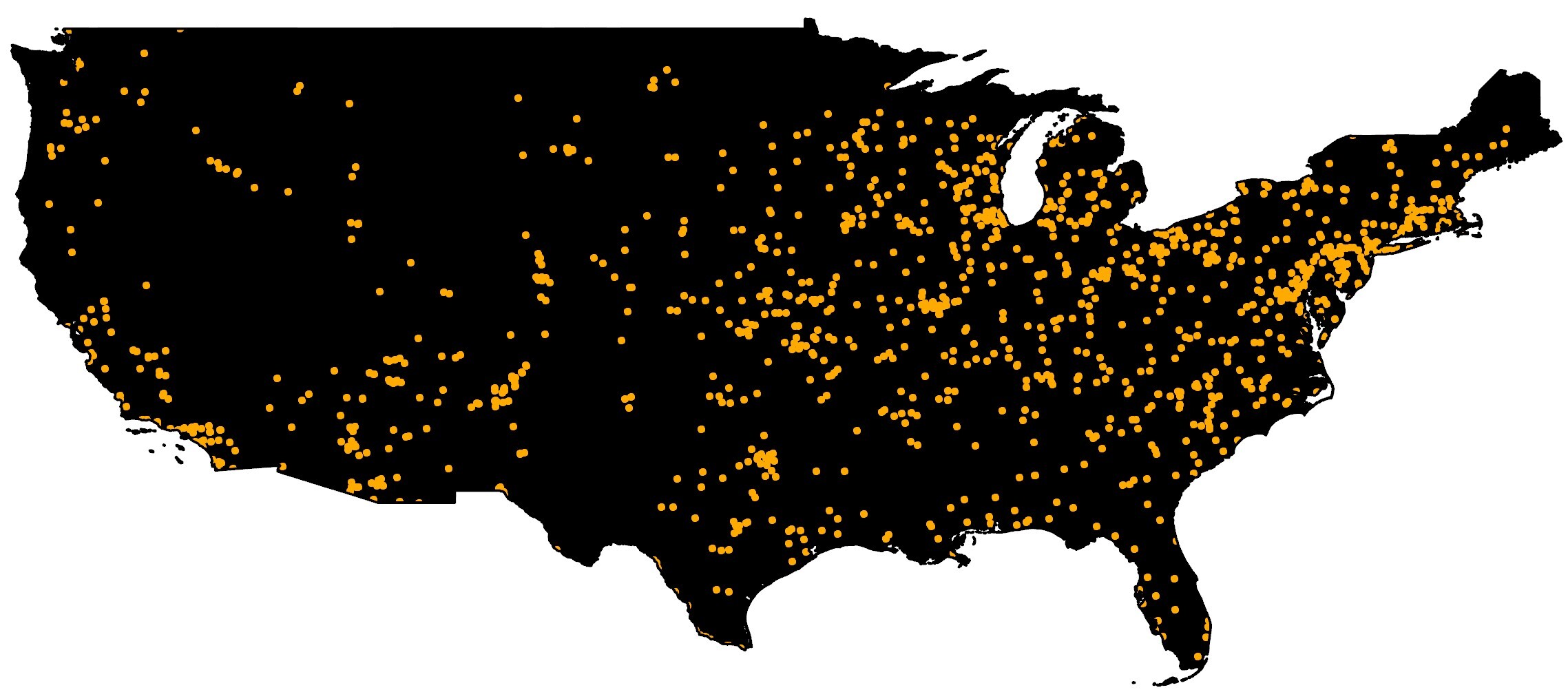}
  \caption{GBIF presence-only}
\end{subfigure}
\caption{Range map of \textit{Abies balsamea} using a query ground-level image and combination of various modalities across the USA.}
\label{fig:sdm_v2}
\end{figure*}

Here we provide additional cross-modal retrieval results of TaxaBind on our TaxaBench-8k dataset (Table~\ref{tab:retr_v2}). It is observed that the retrieval performance improves when embeddings from different modalities are added together. We also present six examples of ground-level image-to-satellite image retrieval (Figure~\ref{fig:sp2sat_v2}). For each example, we present the top-4 most similar satellite images retrieved by our model. We also present range map predicted by our model for \textit{Abies balsamea} in Figure~\ref{fig:sdm_v2}. 

\begin{table*}[!ht]
  \begin{center}
  \begin{tabular}{lcccc}
    \toprule
    Method & Modality & R@1 & R@5 & R@10\\
    \midrule
    \textit{Random Baseline}&-&0.01&0.05&0.11\\
    \midrule
     \multirow{7}{*}{TaxaBind}&\satChar{}\arrowChar{}\imageChar{}&1.87&7.23&12.02\\
    &\imageChar{}\arrowChar{}\satChar{}&1.34&5.42&9.26\\
    &\audioChar{}\arrowChar{}\imageChar{}&1.00&4.10&7.75\\
    &\imageChar{}\plusChar\textChar\arrowChar{}\satChar{}&2.03&8.16&13.66\\
    &\imageChar{}\plusChar\textChar\arrowChar{}\satChar{}\plusChar\locChar&2.39&8.80&14.74\\

    &\envChar{}\arrowChar{}\satChar{}&2.22&11.26&19.40\\
    &\envChar{}\arrowChar{}\satChar{}\plusChar\locChar&3.25&14.18&23.52\\
    \bottomrule
    \end{tabular}
  \caption{Additional retrieval results on our TaxaBench-8k dataset.}
  \label{tab:retr_v2}
  \end{center}
\end{table*}
\begin{figure*}[!t]
\centering
\begin{subfigure}{0.331\linewidth}
  \centering
  \includegraphics[width=\columnwidth]{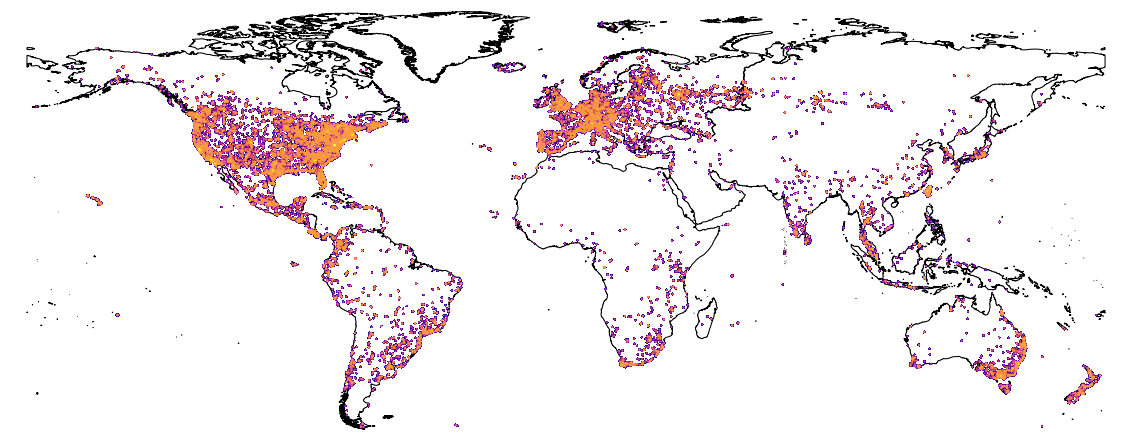}
  \caption{Train}
\end{subfigure}%
\begin{subfigure}{.331\linewidth}
  \centering
  \includegraphics[width=\columnwidth]{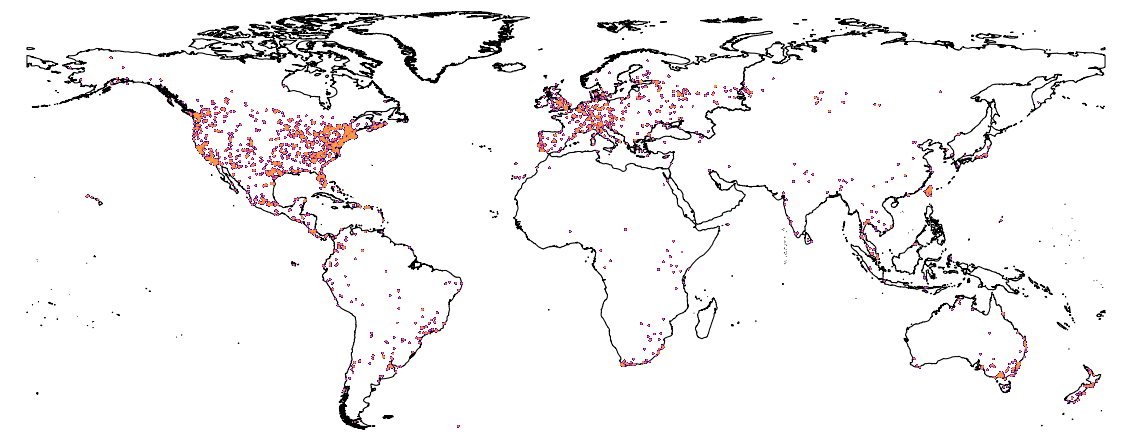}
  \caption{Validation}
\end{subfigure}
\begin{subfigure}{.331\linewidth}
  \centering
  \includegraphics[width=\columnwidth]{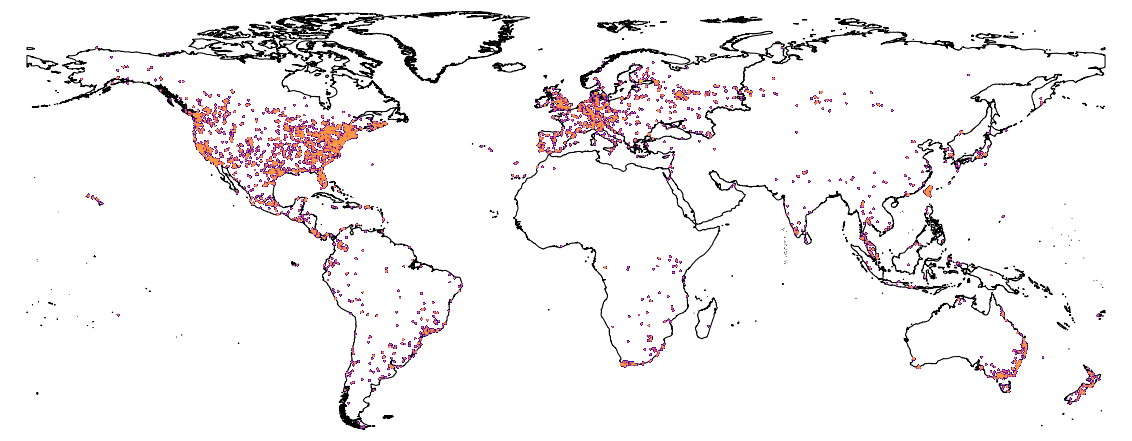}
  \caption{Taxabench-8k}
  \label{taxabench}
\end{subfigure}
\caption{Geographic locations of samples in the training, validation and testing splits of iSoundNat.}
\label{fig:dist}
\end{figure*}
\section{Datasets}
Here we provide additional details about the datasets used for training and evaluating our models.
\subsection{Training Datasets}
\textbf{iSatNat}. This dataset was built by collecting Sentinel-2 Level 2A imagery corresponding to each geolocation in the iNaturalist-2021 dataset. We used the training and validation splits of the dataset and dropped the testing split since it lacked the ground-truth labels. We used the validation split as the unseen testing set. We created a 90:10 split of the training dataset to create the final training and validation sets. Lastly, we applied a minimal filter to remove all the samples lacking geolocation entry.\\ 
\textbf{iSoundNat}. Using the iNaturalist platform, we filtered all the observations with audio recordings and ground-level images to date. We then removed all corrupted audio recordings and converted them to a common format (m4a). This resulted in a total of 88,130 observations. We then used a stratified sampling technique to split the dataset into 85:5:10 (train, validation, test) ratio. The spatial distribution of the dataset is shown in Figure~\ref{fig:dist}.\\
\textbf{WorldClim 2.1}. This dataset consists of 19 bioclimatic variables and an additional elevation map. All the channels are scaled to a resolution of 5 arc minutes.

\section{Evaluation Datasets}
\textbf{TaxaBench-8k}. We extended the testing split of iSoundNat by downloading Sentinel-2 Level 2A imagery corresponding to each location in the split. This dataset is used for evaluating our models on zero-shot image classification and cross-modal retrieval. The spatial distribution of the dataset is shown in Figure~\ref{taxabench}.\\
\textbf{Birds525}~\cite{Piosenka_2023}. This dataset consists of images of bird species across 525 categories. Each image features a single bird species. To evaluate our models, we use the testing split of the dataset which consists of 2,625 samples.\\
\textbf{CUB-200-2011}~\cite{reed2016learning}. This dataset consists of images of 200 bird species. We use the testing split of the dataset which contains 5,794 images.\\
\textbf{BioCLIP-Rare}~\cite{stevens2024bioclip}. This dataset was used for the evaluation of BioCLIP. It contains 400 species categories not present in the TreeOfLife-10M dataset.\\
\textbf{BirdCLEF-2022}~\cite{kahl2022overview}. This dataset contains audio recordings of rare bird species in Hawaii. We use the training split of the dataset which contains annotations. In total, there are 14,852 audio recordings across 141 categories. We use stratified sampling to split the dataset into 85:5:10 (train, validation, test) ratios.\\
\textbf{BirdCLEF-2023}~\cite{kahl2023overview}. This dataset contains audio recordings of bird species in Kenya. We use the training split of the dataset which contains annotations. In total, there are 16,941 audio recordings across 264 categories. We use stratified sampling to split the dataset into 85:5:10 (train, validation, test) ratios.\\
\textbf{BirdCLEF-2024}~\cite{birdclef-2024}. This dataset contains audio recordings of bird species in Western Ghats, India. We use the training split of the dataset which contains annotations. In total, there are 24,459 audio recordings across 182 categories. We use stratified sampling to split the dataset into 85:5:10 (train, validation, test) ratios.\\
\textbf{Ecoregions}. We use the ecoregion map from~\cite{dinerstein2017ecoregion}. The map consists of 846 distinct categories of ecoregions. We randomly sample 100k points around the globe. We split the points into 85:5:10 (train, validation, test) ratio. Then we perform linear probing on our model and report top-1 classification accuracy on the testing split.\\
\textbf{Biome}. We use the biome map from~\cite{dinerstein2017ecoregion}. The map consists of 14 distinct categories of biomes. We randomly sample 100k points around the globe. We split the points into 85:5:10 (train, validation, test) ratio. Then we perform linear probing on our model and report top-1 classification accuracy on the testing split.\\
\textbf{GeoPlant}~\cite{picek2024geoplant}. This dataset consists of the presence-absence of plant species across different countries in Europe. The dataset additionally contains presence-only observations from GBIF. We use only the presence-absence split which contains 88,783 unique locations. We use stratified sampling using country labels to split the set into 85:5:10 (train, validation, test) ratios. For each location, the dataset contains the presence of multiple species. We use SatBird's~\cite{teng2024satbird} training procedure and predict the presence-absence of species at each of the locations.\\
\textbf{SatBird}~\cite{teng2024satbird}. This dataset contains the presence-absence checklist of bird species in two regions: the USA and Kenya. The dataset for the USA is further divided into two seasons: summer and winter. Each location in the dataset is associated with a multispectral Sentinel-2 satellite image.%

\section{Ethics and Limitations}
The models built are a proof-of-concept for demonstrating the benefits of combining multiple modalities for solving ecological problems. Care must be taken when utilizing our models for real-life applications. Additional validation may be necessary before deploying our models. We recognize that the datasets used for training and evaluation may have some spatial bias. However, the goal of this work is not to specifically tackle the issue of spatial bias, but rather to utilize and understand patterns in multiple modalities. We observe that incorporating additional modalities into the framework helps to implicitly account for this bias in the data.

\end{document}